\documentclass[a4paper]{llncs}

\usepackage{amssymb}
\usepackage{graphicx}
\usepackage{amsmath}
\usepackage[dvipsnames]{xcolor}
\usepackage{multicol}
\usepackage{booktabs}
\usepackage{wrapfig}

\usepackage{caption}

\usepackage{url}
\urldef{\mails}\path|{maria.kolos, e.burnaev, a.artemov}@skoltech.ru, anton.marin@skolkovotech.ru|

\usepackage{lineno}

\usepackage{xspace}
\makeatletter
\DeclareRobustCommand\onedot{\futurelet\@let@token\@onedot}
\def\@onedot{\ifx\@let@token.\else.\null\fi\xspace}

\def\eg{\emph{e.g}\onedot} 
\def\ie{\emph{i.e}\onedot} 
\def\cf{\emph{cf}\onedot} 
\def\etc{\emph{etc}\onedot}

\newcommand{\cm}[0]{\checkmark}

\makeatother

\begin{document}

\title{Procedural Synthesis of Remote Sensing Images\\
for Robust Change Detection with Neural Networks}
\author{Maria Kolos$^2$, Anton Marin$^2$, Alexey Artemov$^1$ and Evgeny Burnaev$^1$\thanks{The work was supported by The Ministry of Education and Science of Russian Federation, grant No. 14.615.21.0004, grant code: RFMEFI61518X0004.}}
\institute{Skolkovo Institute of Science and Technology, $^1$ADASE and $^2$Aeronet groups\\
\mails}

\maketitle

\begin{abstract}
Data-driven methods such as convolutional neural networks (CNNs) are known to deliver state-of-the-art performance on image recognition tasks when the training data are abundant. However, in some instances, such as change detection in remote sensing images, annotated data cannot be obtained in sufficient quantities. In this work, we propose a simple and efficient method for creating realistic targeted synthetic datasets in the remote sensing domain, leveraging the opportunities offered by game development engines. We provide a description of the pipeline for procedural geometry generation and rendering as well as an evaluation of the efficiency of produced datasets in a change detection scenario. Our evaluations demonstrate that our pipeline helps to improve the performance and convergence of deep learning models when the amount of real-world data is severely limited. 


\keywords{Remote Sensing, Deep Learning, Synthetic Imagery}
\end{abstract}

\section{Introduction}
\label{sec:introduction}

Remote sensing data is utilized in a broad range of industrial applications including emergency mapping, deforestation, and wildfire monitoring, detection of illegal construction and urban growth tracking. Processing large volumes of the remote sensing imagery along with handling its high variability (\eg, diverse weather/lighting conditions, imaging equipment) provides a strong motivation for developing automated and robust approaches to reduce labor costs.

Recently, data-driven approaches such as deep convolutional neural networks (CNNs) have seen impressive progress on a number of vision tasks, including semantic segmentation, object detection, and change detection~\cite{cheng2017remote,chen2017deeplab,buslaev2018fully,he2017mask,lin2018focal,lin2017feature,alcantarilla2018street}. Such methods offer promising tools for remote sensing applications as they can achieve high performance by leveraging the diversity of the available imagery~\cite{bai2018towards,daudt2018urban,kemker2017deep}. However, in order to successfully operate, most data-driven methods require large amounts of high-quality annotated data~\cite{Gaidon2018,kemker2017deep,gaidon2016virtual,dosovitskiy2017carla}. Obtaining such data in the context of remote sensing poses a significant challenge, as (1)~aerial imagery data are expensive, (2)~collection of raw data with satisfactory coverage and diversity is laborious, costly and error-prone, as is (3)~manual image annotation; (4)~moreover, in some instances such as change detection, the cost of collecting a representative number of rarely occurring cases can be prohibitively high. Unsurprisingly, despite public real-world annotated remote sensing datasets exist~\cite{novikov2018satellite,anniballe2018earthquake,fujita2017damage,daudt2018urban,ChangeDetectionRS2019}, these challenges have kept them limited in size, compared to general-purpose vision datasets such as the ImageNet~\cite{deng2009imagenet}. 

The alternatives considered to avoid dataset collection issues suggest producing synthetic annotated images with the aid of game development software such as \emph{Unity}~\cite{unity3d}, \emph{Unreal Engine 4}~\cite{ue4}, and \emph{CRYENGINE}~\cite{cryengine}. This approach has been demonstrated to improve the performance of computer vision algorithms in some instances~\cite{chen2015deepdriving,muller2018sim4cv,Gaidon2018,qiu2016unrealcv,dosovitskiy2017carla,richter2016playing,shah2018airsim}. Its attractive benefits include (1) flexibility in scene composition, addressing class imbalance issue, (2) pixel-level precise automated annotation, and (3) the possibility to apply transfer learning techniques for subsequent ``fine-tuning'' on real data. However, little work has been done in the direction of using game engines to produce synthetic datasets in the remote sensing domain. Executing procedural changes on large-scale urban scenes is computationally demanding and requires smart optimization of rendering or the object-level reduction (number of polygons, textures quality). Levels of realism rely heavily on the amount of labor on scene design and optimization. Thus, research on the procedural construction of synthetic datasets would contribute to the wider adoption of data-driven methods in the remote sensing domain.

In this work, we focus on the task of change detection, however, it is straightforward to adapt our method to other tasks such as semantic segmentation. We leverage game development tools to implement a semi-automated pipeline for procedural generation of realistic synthetic data. Our approach uses publicly available cartographic data and produces realistic 3D scenes of real territory (\eg, relief, buildings). These scenes are rendered using \emph{Unity} engine to produce high-resolution synthetic RGB images. Taking advantage of real cartographic data and emulation of image acquisition conditions, we create a large and diverse dataset with a low \emph{simulated-to-real} shift, which allows us to efficiently apply deep learning methods. We validate our data generation pipeline on the change detection task using a state-of-the-art deep CNN. 
We observe consistent improvements in performance and convergence of our models on this task with our synthetic data, compared to when using (scarce) real-world data only. 

In summary, our contributions in this work are:
\begin{itemize}
    \item We describe a semi-automatic pipeline for procedural generation of realistic synthetic images for change detection in the domain of remote sensing.
    
    \item We demonstrate the benefits of large volumes of targeted synthetic images for generalization ability of CNN-based change detection models using extensive experiments and a real-world evaluation dataset.
\end{itemize}

The rest of this paper is organized as follows.
In Section~\ref{sec:related}, we review prior work on change detection in the remote sensing domain, including existing image datasets, data generation tools, computational models, and transfer learning techniques. Section~\ref{sec:methods} presents our data generation pipeline. Section~\ref{sec:experiments} poses three experiments investigating the possible benefits of our approach and presents their results. We conclude with a discussion of our results in Section~\ref{sec:conclusion}.

\section{Related work}
\label{sec:related}

\subsection{Computational models for~change detection}
\label{subsec:comp_models}

Change detection in multi-temporal remote sensing images has attracted considerable interest in the research community, where approaches have been proposed involving anomaly detection on time series and spectral indices~\cite{cai2015detecting}, Markov Random Fields and global optimization on graphs~\cite{vakalopoulou2015simultaneous,yu2017change,gu2017change}, object-based segmentation followed by changes classification~\cite{jianya2008review,huang2015automatic,vittek2014land}, and Multivariate Alteration Detection~\cite{wang2017image,jabari2016building} (cf.~\cite{tewkesbury2015critical} for a broader review). These approaches generally work with low-resolution imagery (\eg, 250--500\,m/pixel) and require manual tuning of dozens of hyperparameters to handle variations in data such as sensor model, seasonal variations, image resolution, and calibration. 

Recently, deep learning and CNNs have been extensively studied for classification, segmentation, and object detection in remote sensing images~\cite{zhang2016deep,bruzzone2014review}. However, only a handful of CNN-based change detection approaches exist. Due to the lack of training data~\cite{el2017zoom,saha2018unsupervised} use ImageNet pre-trained models to extract deep features and use super-pixel segmentation algorithms to perform change detection. We only study the influence of pre-training on ImageNet in one of the experiments; otherwise, we train our deep CNN from scratch using our synthetic dataset.
\cite{el2016convolutional,fujita2017damage} proposed a CNN-based method for binary classification of changes given a pair of two high-resolution satellite images.
In contrast, we focus on predicting a dense mask of changes from the two registered images. The closest to our work are~\cite{wiratama2018dual,bai2018towards}, which predict pixel-level mask of changes from the two given images; additionally,~\cite{bai2018towards} uses a U-Net-like architecture as we do. Nevertheless, their models are different from ours, which is inspired by~\cite{buslaev2018fully}.

Due to the scarcity of the available data in some instances, transfer learning techniques have been extensively investigated in many image analysis tasks, including image classification~\cite{yosinski2014transferable,yao2010boosting}, similarity ranking~\cite{zhang2018perceptual} and retrieval~\cite{sharif2014cnn,babenko2014neural}. Additionally, transfer from models pre-trained on RGB images to a more specialized domain, such as magnetic resonance images or multi-spectral satellite images, has been studied for automated medical image diagnostics~\cite{van2015off,MedTransfer,lee2018explainable} and remote sensing image segmentation~\cite{iglovikov2018ternausnet}. In the context of the present work, of particular interest is the transfer learning from synthetic data to real-world data, that has proven effective for a wide range of tasks~\cite{chen2015deepdriving,muller2018sim4cv,Gaidon2018,qiu2016unrealcv,dosovitskiy2017carla,richter2016playing,shah2018airsim}.
In the remote sensing domain, however, synthetic data has been only employed in the context of semantic segmentation~\cite{kemker2017deep}. However, their data generation method relies on pre-created scene geometry, while our system generates geometry based on the requested map data.

\subsection{Image datasets for~change detection}
\label{subsec:related}


\subsubsection{Real-world datasets. }

%
Datasets for change detection are commonly structured in pairs of registered images of the same territory, made in distinct moments in time, accompanied by image masks per each of the annotated changes. With the primary application being emergency mapping, most datasets typically feature binary masks annotating damaged structures across the mapped areas~\cite{fujita2017damage,novikov2018satellite,bourdis2011constrained,anniballe2018earthquake}.
L'Aquila 2009 earthquake dataset~\cite{anniballe2018earthquake} contains data spanning $1.5\times1.5$\,km$^2$ annotated with masks of damaged buildings during the 2009 earthquake. California wildfires~\cite{novikov2018satellite} contains $2.5\times2.5$\,km$^2$ and $5\times8$\,km$^2$ images, representing changes after a 2017 wildfire in California, annotated with masks of burnt buildings. ABCD dataset~\cite{fujita2017damage} is composed of patches for 66\,km$^2$ of tsunami-affected areas, built to identify whether buildings have been washed away by the tsunami. Besides, OSCD dataset~\cite{daudt2018urban} addresses the issue of detecting changes between satellite images from different dates, containing 24~pairs of images from different locations, annotated with pixel-level masks of new buildings and roads. All these datasets are low in diversity and volume, while only providing annotations for a limited class of changes (\eg, urban changes). In contrast, our pipeline can produce massive amounts of highly diverse synthetic images with flexible annotation, specified by the user. Other known datasets, such as the Landsat ETM/TM datasets~\cite{goward2001landsat}, are of higher volume, but have an extremely low spatial resolution (on the order of 10\,m), while featuring no annotation.


\subsubsection{Synthetic datasets and visual modelling tools. }

To the best of our knowledge, AICD dataset~\cite{bourdis2011constrained} is the only published synthetic dataset on change detection in remote sensing domain. It consists of 1000 pairs of $800\times 600$ images. It is a synthetic dataset in which the images are generated using a realistic rendering engine of a computer game, with ground truth generated automatically. The drawbacks of this dataset are low diversity in target and environmental changes and low graphics quality. 



Despite tools for creating synthetic datasets are actively developed and studied in the domain of computer vision~\cite{unity3d,qiu2016unrealcv,ue4}, research on the targeted generation of synthetic datasets for remote sensing applications is still in its infancy. Modern urban modeling packages require manual creation of assets and laborious tuning of rendering parameters (\eg, lighting)~\cite{cityengine,OSM2xp,WorldMachine}. This could be improved by leveraging extensive opportunities offered by game development engines, that combine off-the-shelf professional rendering presets, realistic shaders, and rich scripting engines for fine customization. In our pipeline, procedural generation of geometry and textures is followed by a rendering script, leveraging rich rendering opportunities. Other tools such as DIRSIG~\cite{goodenough2012dirsig} allow simulating realistic multi-spectral renders of the scenes but rely on the existing geometry. Our pipeline, in contrast, enables us to create both geometry-based on real-world map data and realistic renders using a game engine.


\section{Synthesis of territory-specific remote sensing images}
\label{sec:methods}

\subsection{Data requirements and the design choices of our pipeline}
\label{methods:requirements_tools}

A good change detection dataset should contain application-specific target changes, such as, \eg, deforestation or illegal construction, as well as high variability, which can be viewed as non-target changes. Such variability in data commonly involves appearance changes, \eg, lighting and viewpoint variations, diverse directions of shadows, and random changes of scene objects. However, implementing an exhaustive list of non-target changes is too laborious. Thus, we restricted ourselves to the following \emph{general requirements} to the synthetic data:

\begin{itemize}  
    \item \textit{Visual scene similarity.} To reduce the omnipresent \emph{simulated to real} shift, it is necessary that the modeled scenes have a high visual resemblance to the actual scenes. We approach the target territory modeling task by imitating the visual appearance of structures and environment. 

    \item \textit{Scale.} To match the largest known datasets in spatial scale, we have chosen to model scenes with large spatial size. 
    In our dataset, scenes are generated with spatial extents of square kilometers, a spatial resolution of less than 1\,m, and image resolution of tens of Megapixels.
    
    \item \textit{Target changes.} We have chosen to only model damaged buildings as target class as they are of general interest in applications such as emergency mapping (see, \eg, \cite{jianya2008review,jabari2016building,tewkesbury2015critical,saha2018unsupervised,el2017zoom,wiratama2018dual,bai2018towards,el2016convolutional,vittek2014land,huang2015automatic}). They are also straightforward to implement in our pipeline with procedural geometry generation and rendering scripts.
    
    \item \textit{Viewpoint variations}. Real-world multi-temporal remote sensing images for the same territory are commonly acquired using varying devices (\eg, devices with different field of view) and viewpoints. The acquisition is commonly performed at angles not exceeding 25~deg., and the data are then post-processed by registration and geometric correction. In our pipeline, we imitate the precession of a real satellite by randomly changing the image acquisition angle and the field of view.
    
    \item \textit{Scene lighting changes}. Scene illumination is commonly considered to consist of a point-source illumination (produced by the Sun) and an ambient illumination due to the atmospheric scattering of the solar rays. In our work, we consider the changes in the Sun's declination angle and model both components of the illumination.
 
    \item \textit{Shadows}. Real-world objects cast shadows that are irrelevant variations and should be ignored. We model realistic shadows again by varying the Sun's declination angle. 

\end{itemize}

\begingroup

\begin{figure}[t]
\begin{center}
\includegraphics[width=0.78\columnwidth]{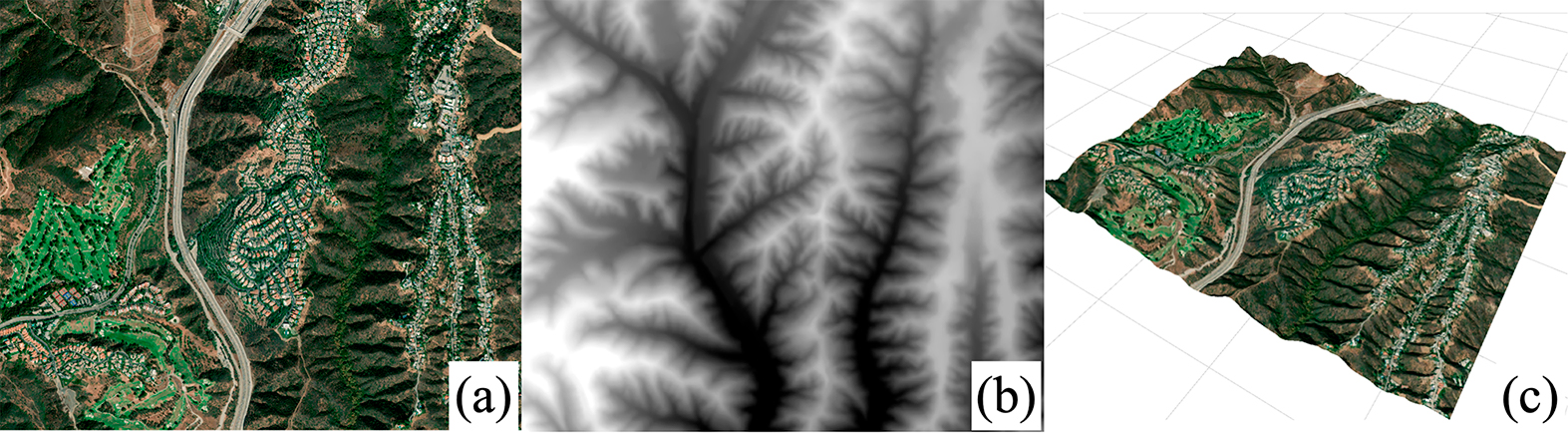}
\caption{Terrain reconstruction: (a) satellite imagery, (b) altitude data, (c) 3d model.}
\label{fig:terrain_reconstruction_texturing}
\bigskip
\includegraphics[width=0.78\columnwidth]{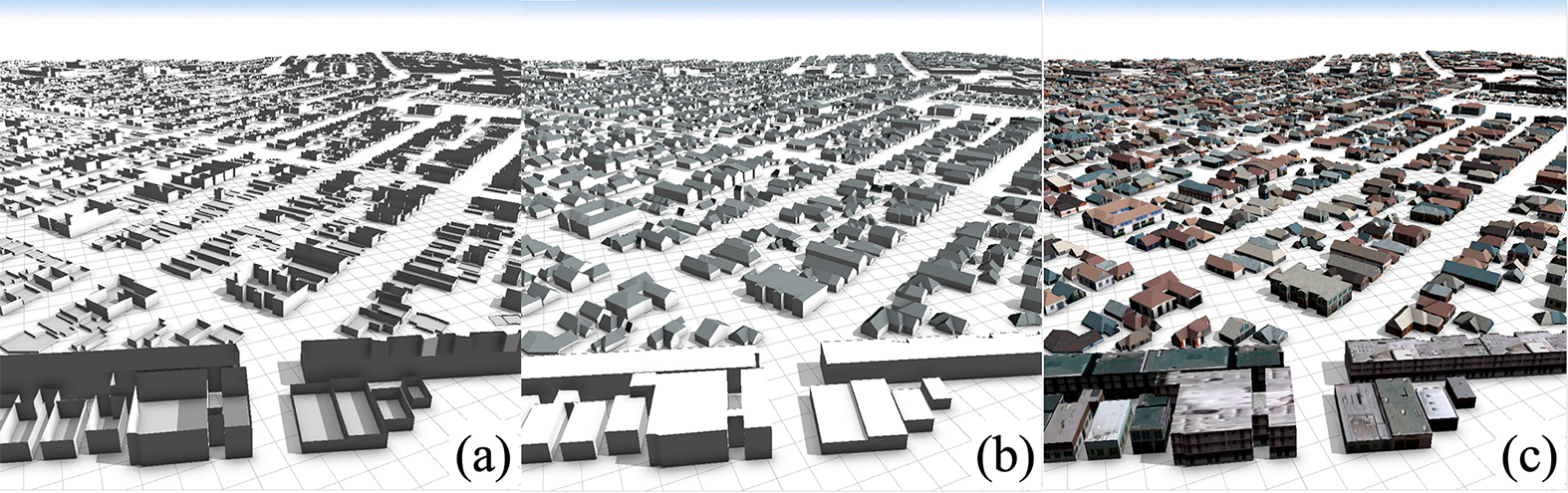}
\caption{Geometry generation: (a) polygon extrusion, (b) adding roofs, (c) texturing.}
\label{fig:extrusion_roofs_texturing}
\end{center}
\end{figure}

\endgroup

To meet these requirements, we have developed a~two-stage pipeline consisting of geometry generation and rendering steps. The entire routine is semi-automatic and involves two widely used 3D engines. Specifically, we use \emph{Esri CityEngine}~\cite{cityengine} to procedurally build geometry from real-world map data and~\emph{Unity}~\cite{unity3d} to implement the logic behind dataset requirements and leverage rendering capabilities. The reasons behind our choice of \emph{CityEngine} as our geometry manipulation tool are its flexibility in the procedural geometric modeling and built-in UV/texturing capabilities. Other tools, commonly implemented as plugins for \emph{Unreal Engine 4} (\eg, \emph{StreetMap}\footnote{\url{https://github.com/ue4plugins/StreetMap}}) and \emph{Unity} (\eg, \emph{Mapbox Unity SDK}\footnote{\url{https://www.mapbox.com/unity}}), offered significantly less freedom. In these tools, either non-textured or textured but oversimplified shapes (\eg, as simple as boxes) are the only objects available for urban geometry generation.

\emph{Unity} game engine was selected to execute a generation procedure of change detection dataset. Compared to CAD software often used for the production of datasets, game engines offer advantages such as powerful lighting/shadows out of the box and scripting possibilities. Additionally, \emph{Unity} allows implementing target changes, controlling lightning and viewpoint changes, and adjusting change rate in the dataset. Certain features in \emph{Unity} are more suited to our needs, compared to \emph{Unreal Engine 4}. For instance, we have found shadows in \emph{Unity} to be more stable while rendering scenes from large distances, and the \emph{Layers} feature\footnote{\url{https://docs.unity3d.com/Manual/Layers.html}} to add more flexibility by allowing to exclude objects from rendering or post-processing. It is natural, however, that we had to execute some initial settings of \emph{Unity} before the generation, as the typical requirements of the remote sensing domain differ from those of 3D games. We describe these settings in Section~\ref{sec:experiments}.

\subsection{Geometry generation}
\label{methods:geometry}

To procedurally generate geometry in \emph{CityEngine}, we obtain cartographic data (vector layers) from \emph{OpenStreetMap} and elevation data (a~geo-referenced GeoTIFF image) from \emph{Esri World Elevation}. Vector data contains information about the geometry of buildings and roads along with semantics of land cover (forests, parks, \etc), while elevation data is used to reconstruct terrain.

First, we reconstruct terrain using the built-in functionality in \emph{CityEngine}, obtaining a textured 3D terrain mesh (see Figure~\ref{fig:terrain_reconstruction_texturing}), to which we $z$-align flat vector objects; second, we run our geometry generation procedure implemented in the engine's rule-based scripting language; last, we apply textures to the generated meshes. Our implementation of the geometry generation involves extruding the polygon of a certain height and selecting a randomly textured roof (architectural patterns such as rooftop shapes are built into \emph{CityEngine}), see Figure~\ref{fig:extrusion_roofs_texturing}. We have experimented with more complex geometry produced by operations such as polygon splitting and repeating; however, such operations (\eg, splitting) increase the number of polygons significantly without adding much to the scene detail, render quality, or performance of change detection models. Focusing on our scalability and flexibility requirements, we avoid overloading our scenes with objects of redundant geometry. 

We select two types of buildings to construct our scenes: small buildings with colored gable roofs and concrete industrial-looking structures with flat roofs, see Figure~\ref{fig:buildings_realism}. A set of roof textures has been selected manually, \emph{CityEngine} built-in packs of textures for \emph{OpenStreetMap} buildings were used for facades. We approach the emergency mapping use-case by texturing buildings footprints to imitate damaged appearance, see Figure~\ref{fig:structures_ashes}.
\begingroup
\begin{figure} 
\vspace*{-0.4cm}
\begin{minipage}[t]{0.49\linewidth}
    \begin{center}
    
    \includegraphics[width=1.0\columnwidth]{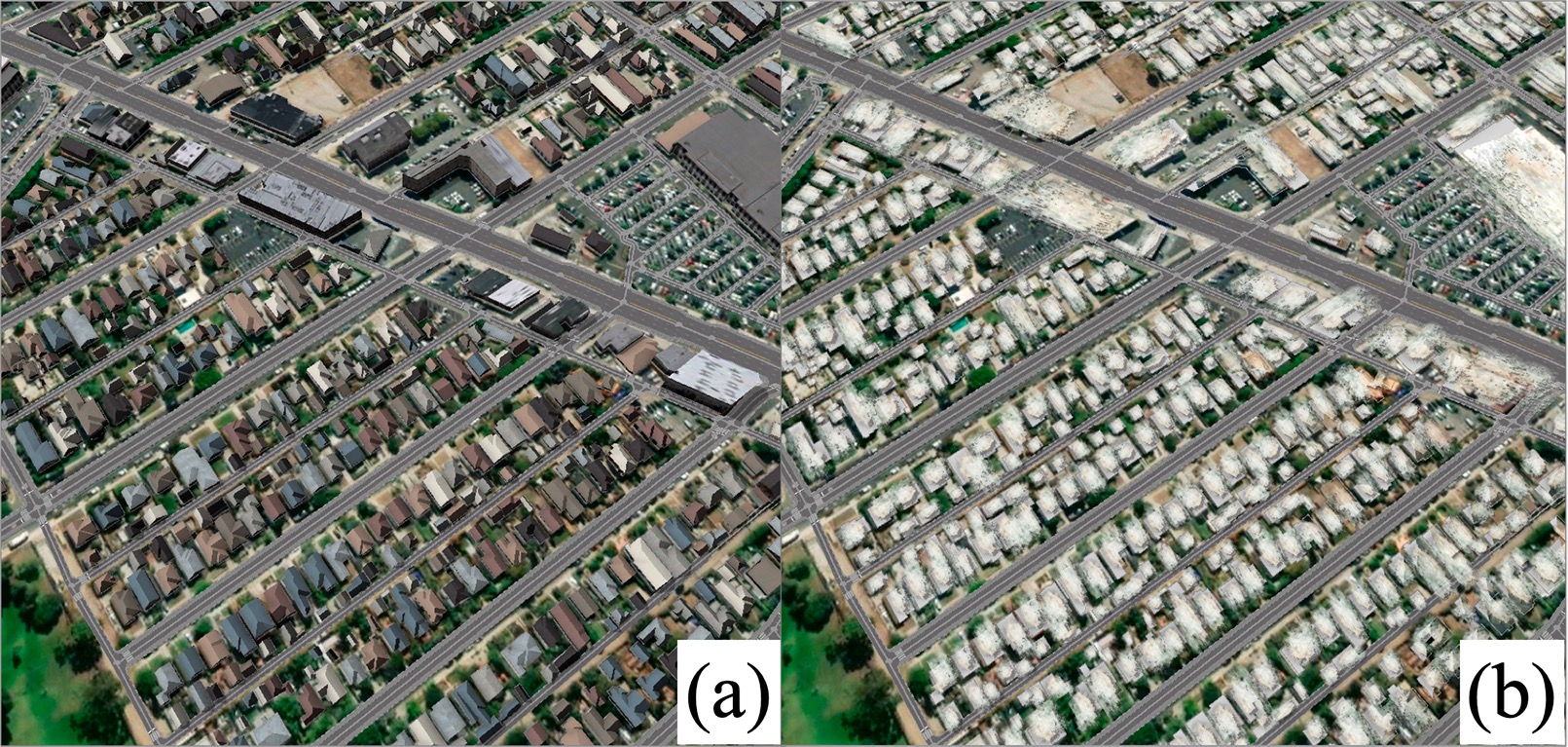}
    \caption{Addressing the case of emergency mapping: scene model (a) before and (b) after the emergency situation (note that some structures are missing).}
    \label{fig:structures_ashes}
    \end{center}
\end{minipage}%
    \hfill%
\begin{minipage}[t]{0.45\linewidth}
    \begin{center}
    \includegraphics[width=0.9\columnwidth]{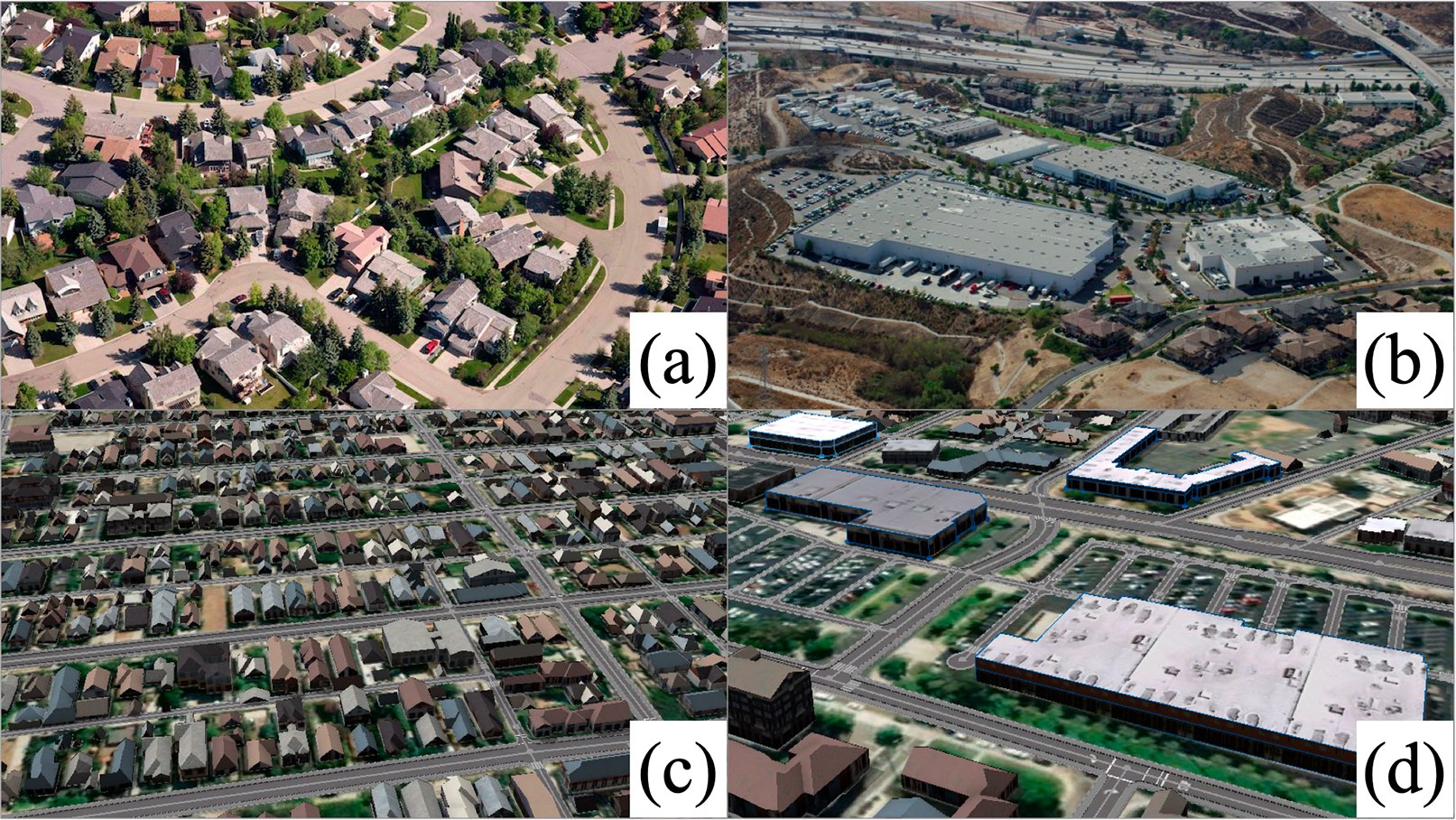}
    \caption{The two models of buildings used in our pipeline: (a) photo and (c) our render of \emph{residential houses}, (b) photo and (d) our render of \emph{industrial structures}.}
    \label{fig:buildings_realism}
    \end{center}
\end{minipage} 
 \vspace*{-0.5cm}
\end{figure}

\endgroup

\subsection{Rendering}
\label{methods:rendering}


We construct the synthetic dataset by rendering the generated geometry using built-in functionality in \emph{Unity}, obtaining high-definition RGB images. 

To achieve a high degree of variations, we leverage rich scripting capabilities in \emph{Unity}, that allow flexible scene manipulation via scripts written in C\#. As shadow casting and lighting algorithms are built in, we only adjust their parameters, as indicated in Section~\ref{experiments:datasets}. We implement target changes by randomly selecting object meshes and placing them onto the separate layer: the camera will not render these meshes, rendering corresponding damaged footprints instead. Non-target variations are further added to scenes modified by target changes, by random alterations of lighting and camera parameters, resulting in $m$ different image acquisition conditions per each target change. 

Each element in the dataset was obtained after three rendering runs: first, we render the original scene; second, we apply both target and non-target changes, moving the changed objects in the separate layer; finally, only the layer with the changed objects is rendered to obtain annotations.

\section{Experiments}
\label{sec:experiments}

We demonstrate the effectiveness of our pipeline in an emergency mapping scenario, where the goal is to perform a rapid localization and assessment of incurred damage with extremely limited amounts of annotated data~\cite{novikov2018satellite}. To this end, we produce a synthetic training dataset in Section~\ref{experiments:datasets} and design a series of experiments to investigate the influence of training strategy on the change detection performance with different data volumes in Sections~\ref{experiments:eval_setup}--\ref{experiments:results}.

\subsection{Datasets}
\label{experiments:datasets}





\subsubsection{California wildfires (CW). }
\label{experiments:datasets_realworld}

The dataset contains high-resolution satellite images depicting cases of wildfires in two areas of Ventura and Santa Rosa counties, California, USA. The annotation has been created manually~\cite{novikov2018satellite}. 

\subsubsection{Synthetic California wildfires (SynCW) dataset. }
\label{experiments:datasets_synthetic}
Using our pipeline, we created a synthetic training dataset for the California wildfires case study. 
We collected \emph{OpenStreetMap} and \emph{ESRI world elevation} data from the area of interest in Ventura and Santa Rosa counties, California, USA. Two major meta-classes imported from \emph{OpenStreetMap} data were \emph{building} (including \emph{apartments, garage, house, industrial, residential, retail, school,} and \emph{warehouse}) and \emph{highway} (road structures including \emph{footway, residential, secondary, service,} and \emph{tertiary}). We generate geometry using our pipeline configured according to Table~\ref{table:generation_pipeine_parameters}.
We add target changes by randomly selecting 30--50\% of \emph{buildings} geometry. 
To introduce non-target changes in the dataset, we select $m=5$ different points of view, positioning camera at zenith and at four other locations determined by an inclination angle $\alpha$ from an axis pointing to zenith (we select $\alpha$ uniformly at random from $[5^\circ, 10^\circ]$), and orienting it to the center of the scene. To model daylight variations, we select Sun's declination angle uniformly at random from $[30^\circ, 140^\circ]$. In the resulting dataset, the generated scenes of 4 distinct locations have spatial extents of approximately $2 \times 2$\,km$^2$ with spatial resolution of 0.6\,m and image resolution of $3072 \times 3072$\,px.

\begingroup

\begin{figure}[t] 
\begin{minipage}[b]{0.55\linewidth}
    \begin{center}
{
\renewcommand{\arraystretch}{0.9}

\begin{tabular}{@{}ll@{}}
\toprule
\multicolumn{2}{l}{Geometry generation parameters \emph{(CityEngine)}}                                                                                                \\ \midrule
\quad Export format      & Autodesk FBX \\
\quad Terrain            & Export all terrain layers \\
\quad Mesh granularity   & One model per start shape \\
\quad Merge normals      & Force separated normals \\
\quad Center             & Checked \\ \midrule
\multicolumn{2}{l}{Rendering parameters \emph{(Unity)}}                                                                                                               \\ \midrule
\quad Shadows            & Hard and soft shadows \\
\quad Shadow distance    & 1100 \\
\quad Shadow resolution  & Very high \\
\quad AntiAliasing       & $8\times$ multi-sampling \\ \midrule
\multicolumn{2}{l}{Camera settings \emph{(Unity)}}                                                                                                                    \\ \midrule
\quad Clipping planes    & 0, 1200 \\
\quad Field of view      & 30.0 \\
\bottomrule
\end{tabular}
\caption{Configuration of geometry generation and rendering parameters in our evaluation.}
\label{table:generation_pipeine_parameters}
}
    
    \end{center}
\end{minipage}%
    \hfill%
\begin{minipage}[b]{0.4\linewidth}
    \begin{center}
    
    \begin{tabular}{@{}p{2cm}cccc@{}}
    \toprule
    Dataset & \rotatebox[origin=c]{90}{Large-scale} & \rotatebox[origin=c]{90}{Real-world} & \rotatebox[origin=c]{90}{Target domain} & \rotatebox[origin=c]{90}{CD annotation} \\ \midrule
    ImageNet~\cite{ImageNet}                & \cm      & \cm       & --        & --        \\
    CW~\cite{novikov2018satellite}          & --       & \cm       & \cm       & \cm       \\
    Our SynCW                               & \cm      & --        & \cm       & \cm       \\
    \bottomrule
    \end{tabular}
    \caption{Datasets considered in our evaluation.}
    \label{table:datasets_summary}  
    
    \end{center}
\end{minipage} 
\end{figure}

\endgroup

\subsection{Our change detection model and training procedure}
\label{experiments:architecture}

\begingroup
\begin{figure}[t]
\begin{center}
\includegraphics[width=1.0\columnwidth]{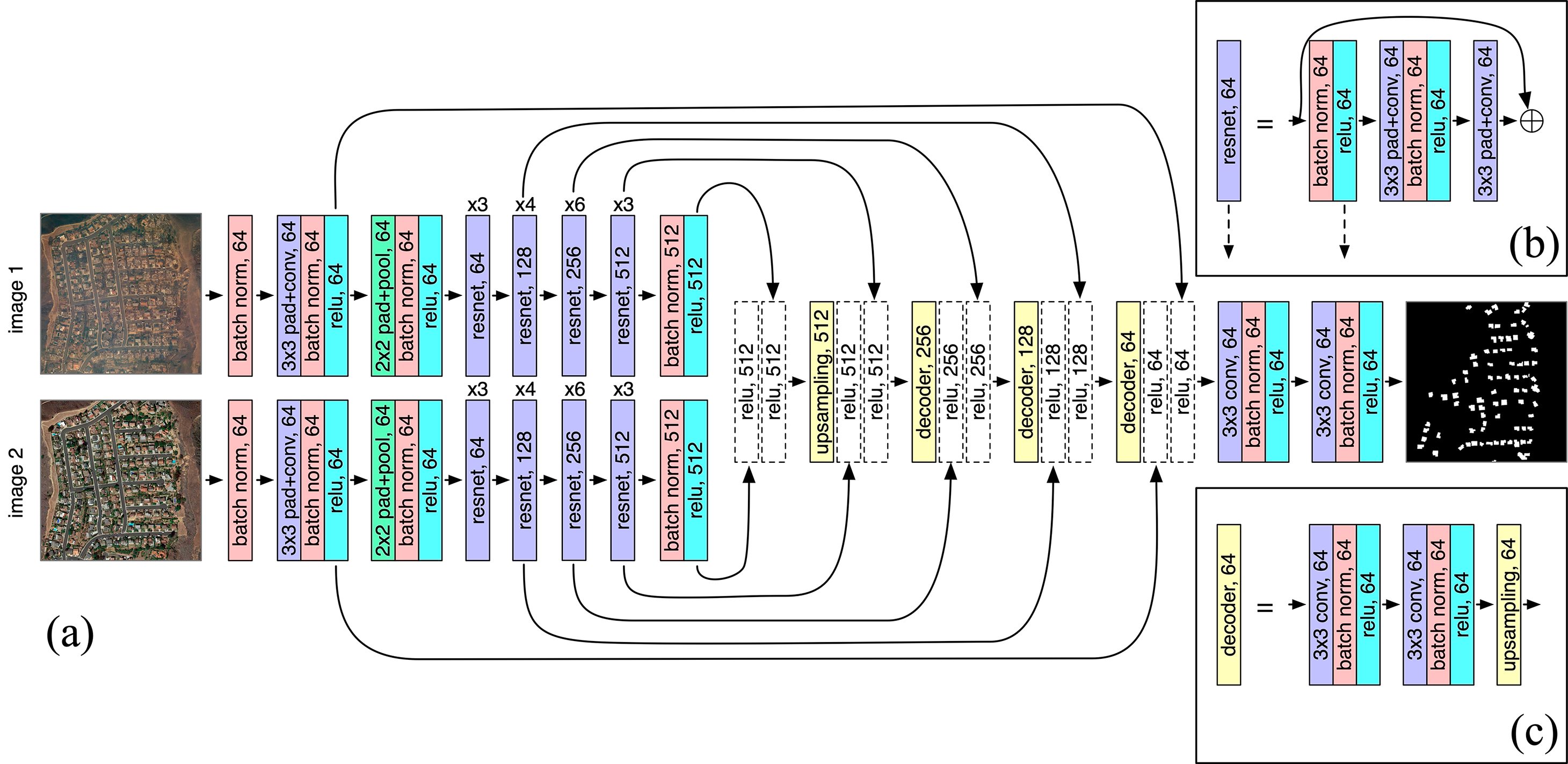}
\caption{Our siamese neural network for change detection: (a) overall architecture, (b) ResNet units (dashed arrows indicate skip-connections, if present), (c) decoder blocks.}
\label{fig:model-architecture}
\end{center}
\end{figure}
\endgroup

When designing our change detection architecture, we take inspiration from recent progress in semantic segmentation~\cite{ronneberger2015u,buslaev2018fully,iglovikov2018ternausnet}. Our architecture is a Siamese U-Net~\cite{ronneberger2015u} with residual units~\cite{he2016deep} in~encoder blocks and upsampling units in~decoder blocks, which can be viewed as a~Siamese version of a~segmentation model from~\cite{buslaev2018fully} (see Figure~\ref{fig:model-architecture}). While our model is composed of well-known building blocks, to the best of our knowledge, we are the first to study its Siamese version in a~change detection setting. While selecting the top-performing architecture is beyond the scope of this paper, we have found our architecture to consistently outperform those examined previously~\cite{ronneberger2015u,iglovikov2018ternausnet} in all settings we have considered.

\begin{figure}
    \begin{center}
        \includegraphics[width=0.58\columnwidth]{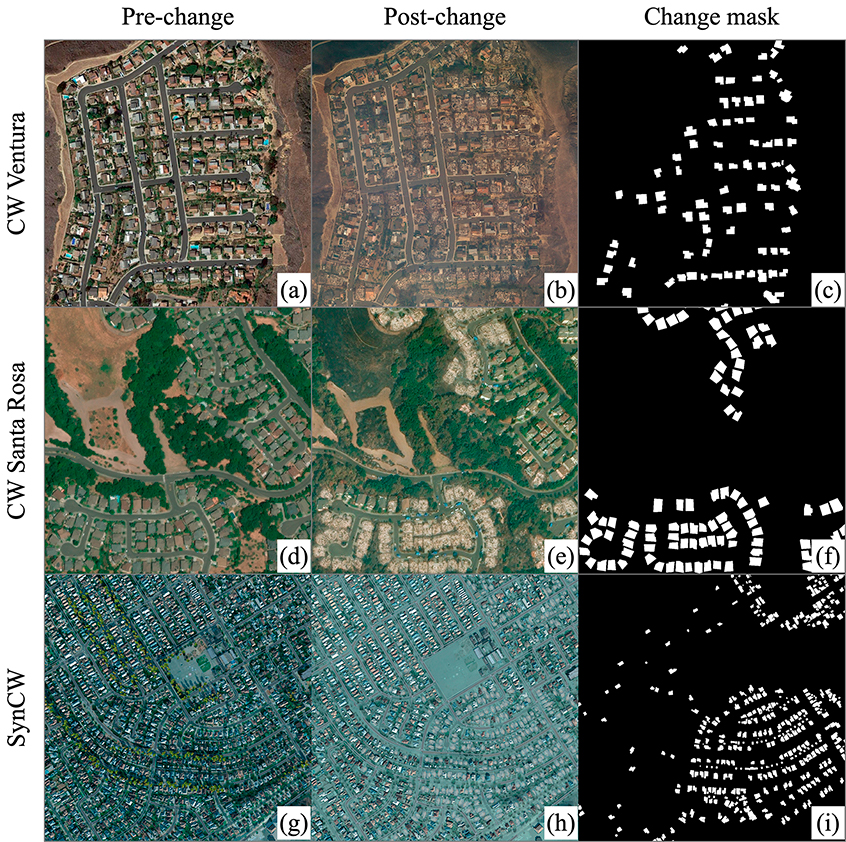}
    \end{center}
    \caption{Visual comparison of our training and evaluation data: (a)--(c) CW, Ventura area, (d)--(f) CW, Santa Rosa area, (g)--(i) synthetic California wildfires (SynCW).}
    \label{fig:ventura_santarosa_syncw}
\end{figure}


To study the benefits of pre-training on a large image dataset, we have kept the encoder architecture a~replica of ResNet-34 architecture~\cite{he2016deep}. In all settings, the models were trained using $352 \times 352$ patches for 20 epochs. Adam optimizer~\cite{kingma:adam} was used with a batch size of~8 and initial learning rate of~$10^{-4}$.

\subsection{Metrics}
\label{experiments:metrics}

To evaluate our models, we chose two performance measures standard for~segmentation tasks: \textit{Intersection over Union} (IoU) and F1-measure, both obtained by applying the threshold 0.5 to the confidence output. Given a pair of binary masks, IoU can be interpreted as a pixel-wise metric that corresponds to localization accuracy between these two samples, $\text{IoU}(A, B) =\frac{\left| A \cap B\right|}{\left| A \cup B\right|} = \frac{\left| A \cap B\right|}{\left| A \right| + \left| B \right| - \left|A \cap B \right|}$. F1-measure is the harmonic mean of precision and recall values between the predicted and ground truth masks: $\text{F1} = 2 \cdot (\text{Precision}^{-1} + \text{Recall}^{-1}).$





\subsection{The evaluation setup}
\label{experiments:eval_setup}


When training a deep learning-based change detection model, an annotated real-world remote sensing dataset (\eg, CW~\cite{novikov2018satellite}), would be a natural choice; however, its volume does not allow training an architecture such as ours from scratch (\ie, starting from randomly-initialized weights). A stronger initialization is commonly obtained with models pre-trained on ImageNet~\cite{ImageNet}, a large-scale and real-world dataset. Unfortunately, ImageNet contains images from a completely different domain; thus, it is unclear whether features trained on ImageNet would generalize well for the change detection scenario. Furthermore, the decoder cannot be initialized and must still be trained. In our setting, SynCW, which is a target-domain large-scale dataset with change annotations, could be employed, but would training on synthetic images lead to good generalization? Thus, there exists no definitive choice of a training data source (\cf Table~\ref{table:datasets_summary}); as we demonstrate further, the choice of training strategy is crucial for achieving high performance.

We design seven training strategies for our task, summarized in Table~\ref{table:results}.
Strategy \emph{A} would be a standard setting with excessive amounts of data. 
In strategies \emph{B-1} and \emph{B-2}, we attempt to model the \emph{synthetic-to-real} transfer scenario.  During pre-training, we either randomly initialize and freeze the encoder (\ie, set its learning rate to zero, \emph{B-1}) or train it (\emph{B-2}). 
Strategies \emph{C}, \emph{D-1}, \emph{D-2}, and \emph{E} all initialize the encoder with ImageNet-pretrained weights, a widely used initialization: \emph{C} realizes a common transfer learning setting, \emph{D-1} and \emph{D-2} proceed in two fine-tuning stages and use the synthetic, then the target training set, either training decoder only (\emph{D-1}, similarly to \emph{B-1}) or the entire model (\emph{D-2}).
\emph{E} is a common transfer learning setting widely used in, \eg, Kaggle\footnote{\url{https://www.kaggle.com}} competitions, leveraging strong augmentations (\eg, rotations, flips, brightness changes, blur, and noise).


Following~\cite{novikov2018satellite}, we use a pair of Ventura train images ($4573 \times 4418$\,px) for training or fine-tuning our models. As our goal is to study the effect of decreasing volumes of real-world data, we crop a random patch from these images, setting the ratio of patch area to the full image area to be 1, 1/2, 1/4, 1/8, and 1/16. A non-overlapping pair of Ventura test images ($1044 \times 1313$\,px) and a pair of visually distinct Santa Rosa images ($2148 \times 2160$\,px) are held out for testing. Note that when testing on Santa Rosa, we do not fine-tune on the same data to test generalization ability. 
In all experiments, we preserve the same architectural and training details as described in Section~\ref{experiments:architecture}. We release the code used to implement and test our models\footnote{\url{https://github.com/mvkolos/siamese-change-detection}}.


{
\renewcommand{\arraystretch}{1.15}

\begin{table}[t]
\centering
\resizebox{\textwidth}{!}{%
\begin{tabular}{@{}llllcccccccc@{}}
\toprule
 & \multicolumn{3}{c}{Training datasets} & \multicolumn{2}{c}{Ventura full} & \multicolumn{2}{c}{Ventura 1/16} & \multicolumn{2}{c}{SR full} & \multicolumn{2}{c}{SR 1/16} \\ 
\cmidrule(l){2-4}
\cmidrule(l){5-6} 
\cmidrule(l){7-8} 
\cmidrule(l){9-10}
\cmidrule(l){11-12}
\multicolumn{1}{c}{} & Init. & Pre & Fine & IoU & F1 & IoU & F1 & IoU & F1 & IoU & F1 \\ \midrule
\textit{A} & \multicolumn{1}{c}{--} & CW & \multicolumn{1}{c}{--} & 0.695 & 0.820 & 0.310 & 0.460 & 0.487 & 0.504 & 0.337 & 0.639 \\
\textit{B-1} & \multicolumn{1}{c}{--} & SynCW$^*$ & CW & 0.713 & 0.832 & 0.312 & 0.476 & 0.487 & 0.549 & 0.392 & 0.563 \\
\textit{B-2} & \multicolumn{1}{c}{--} & SynCW & CW & 0.702 & 0.825 & 0.327 & 0.490 & 0.487 & 0.629 & 0.331 & 0.497 \\
\textit{C} & ImageNet & \multicolumn{1}{c}{--} & CW & 0.716 & 0.835 & 0.499 & 0.704 & 0.626 & 0.770 & 0.435 & 0.607 \\
\textit{D-1} & ImageNet & SynCW$^*$ & CW & 0.714 & 0.833 & \textbf{0.572} & \textbf{0.735} & \textbf{0.680} & \textbf{0.800} & \textbf{0.649} & \textbf{0.787} \\
\textit{D-2} & ImageNet & SynCW & CW & 0.718 & 0.835 & \textbf{0.580} & \textbf{0.744} & \textbf{0.684} & \textbf{0.812} & \textbf{0.631} & \textbf{0.774} \\
\textit{E} & ImageNet & \multicolumn{1}{c}{--} & CW$^{**}$ & \textbf{0.724} & \textbf{0.840} & 0.317 & 0.458 & 0.034 & 0.066 & 0.044 & 0.084 \\ \bottomrule
\end{tabular}%
}
\caption{Statistical change detection results with models trained using strategies \emph{A}--\emph{E}. When using CW as a fine-tuning target, we only select a training subset; (*) indicates frozen encoder, (**) indicates augmentations (see Section~\ref{experiments:eval_setup}).}
\label{table:results}
\end{table}

}

\subsection{Results}
\label{experiments:results}

We present the statistical results of our evaluation in Table~\ref{table:results}. As expected, when training data is present in large volumes (\eg, using augmentations in strategy \emph{E}), models pre-trained on ImageNet perform well. However, when the volume of real-world data supplied for fine-tuning decreases (up to 1//16), such strategies lead to unpredictable results (\eg, for strategy \emph{E} on Ventura test images IoU measure drops by a factor of 2.3 from 0.724 to 0.317). In contrast, fine-tuning using our synthetic images helps to retain a significant part of the efficiency and leads to a more predictable change in the quality of the resulting model (\eg, in strategy \emph{D-1} we observe a decrease in IoU by 20\% only from 0.714 to 0.572). We note how for Santa Rosa images, the performance of models trained without synthetic data degrades severely, while fine-tuning using our synthetic images helps to suffer almost no drop in performance. We plot IoU/F1 vs. volume of used data in Figure~\ref{fig:results_quality_vs_data} to visualize this. We also display qualitative change-detection results in Figure~\ref{fig:visual_results_1_16}. Note how the output change masks tend to become noisy for strategies \emph{A}, \emph{C}, and \emph{E}, and less so for \emph{D-1} and \emph{D-2}. Our synthetic data also leads to faster convergence (see Figure~\ref{fig:results_learning_curves}).

\begingroup
\begin{figure}[t]
\begin{center}
\vspace*{-2cm}
\includegraphics[width=1.0\columnwidth]{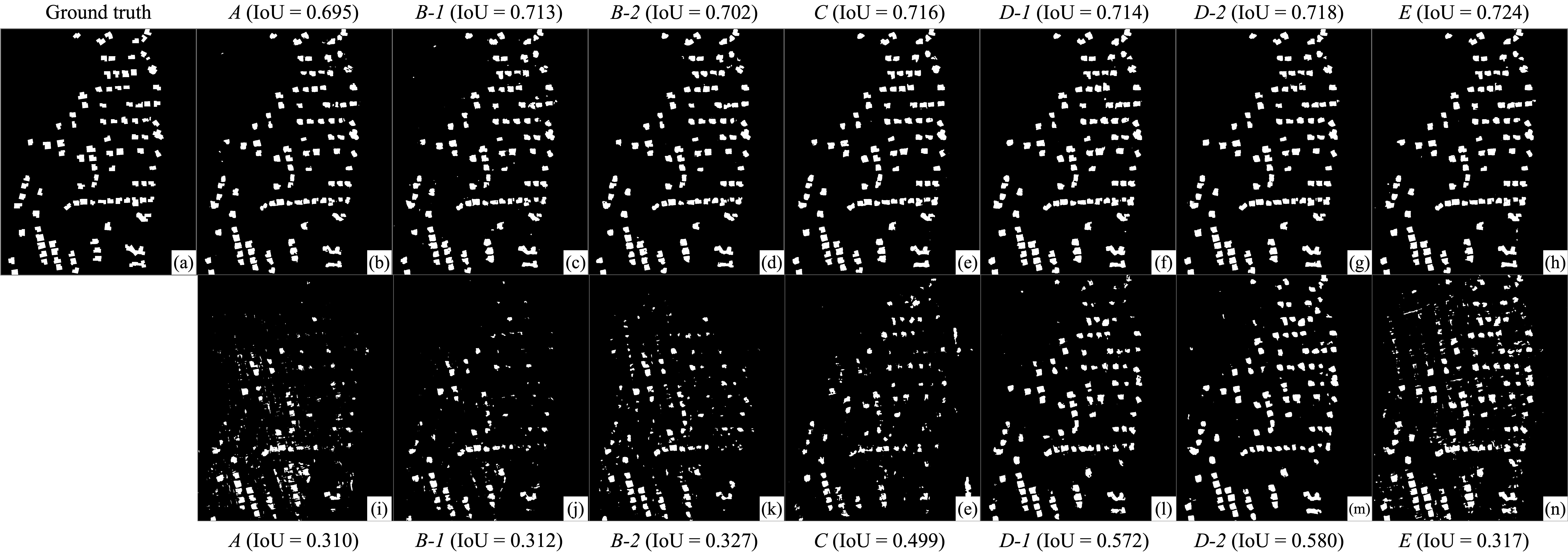}
\caption{Qualitative change detection results: (a) ground truth change mask and (b)--(n): change masks predicted by models \emph{A}--\emph{E}, trained on full real-world dataset (upper row (b)--(h)) and 1/16 of real-world data (lower row (i)--(n)).}
\label{fig:visual_results_1_16}
\end{center}
\end{figure}
\endgroup

\begingroup

\begin{figure} 
\begin{minipage}[t]{0.225\linewidth}
    \begin{center}
    \vspace*{-0.5cm}
    \includegraphics[width=1.0\columnwidth]{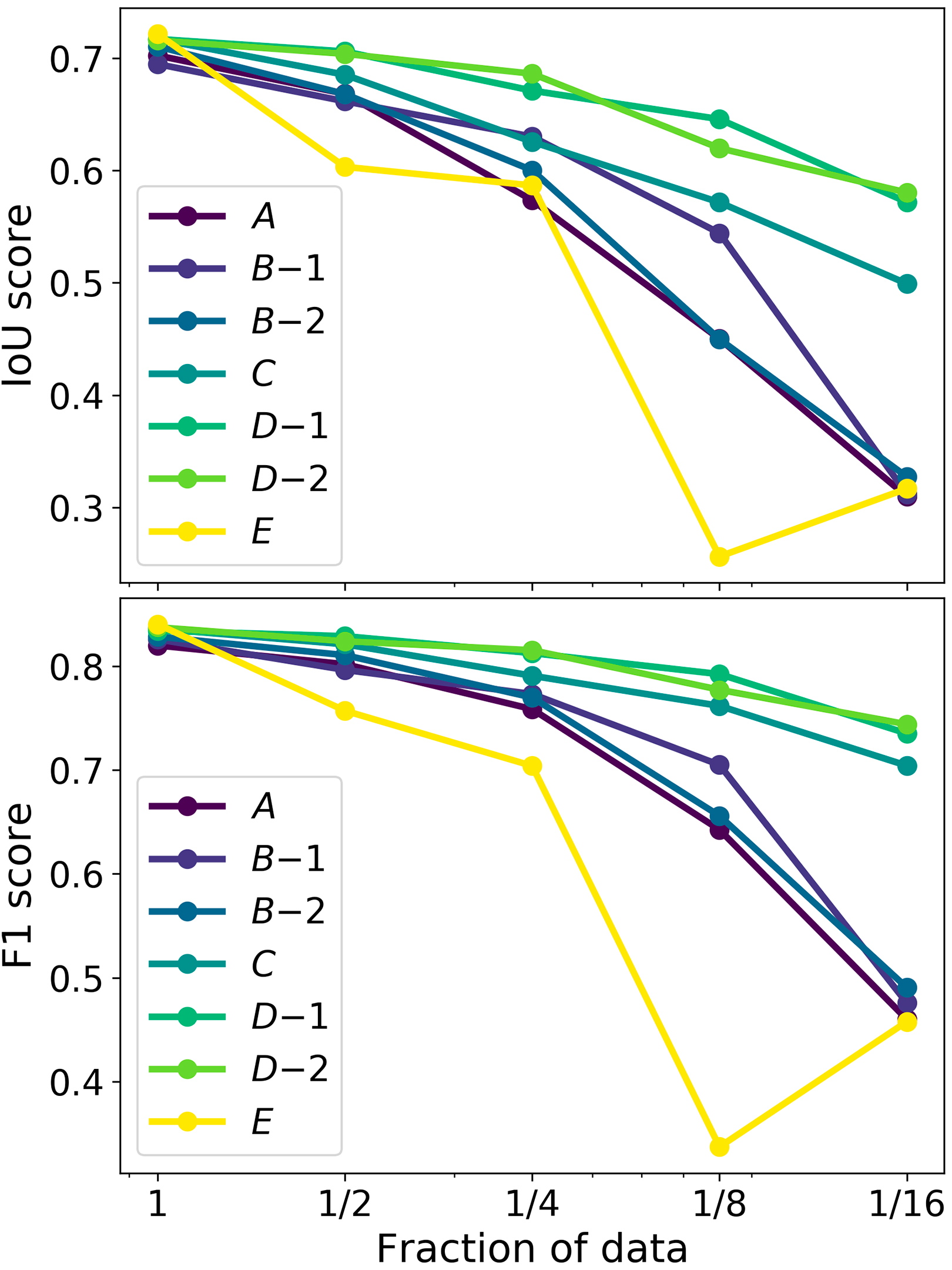}
    \caption{Validation performance when fine-tuning with decreasing volumes of real-world data.}
    \label{fig:results_quality_vs_data}
    \end{center}
\end{minipage}%
    \hfill%
\begin{minipage}[t]{0.725\linewidth}
    \begin{center}
    \vspace*{-0.5cm}
    \includegraphics[width=1.0\columnwidth]{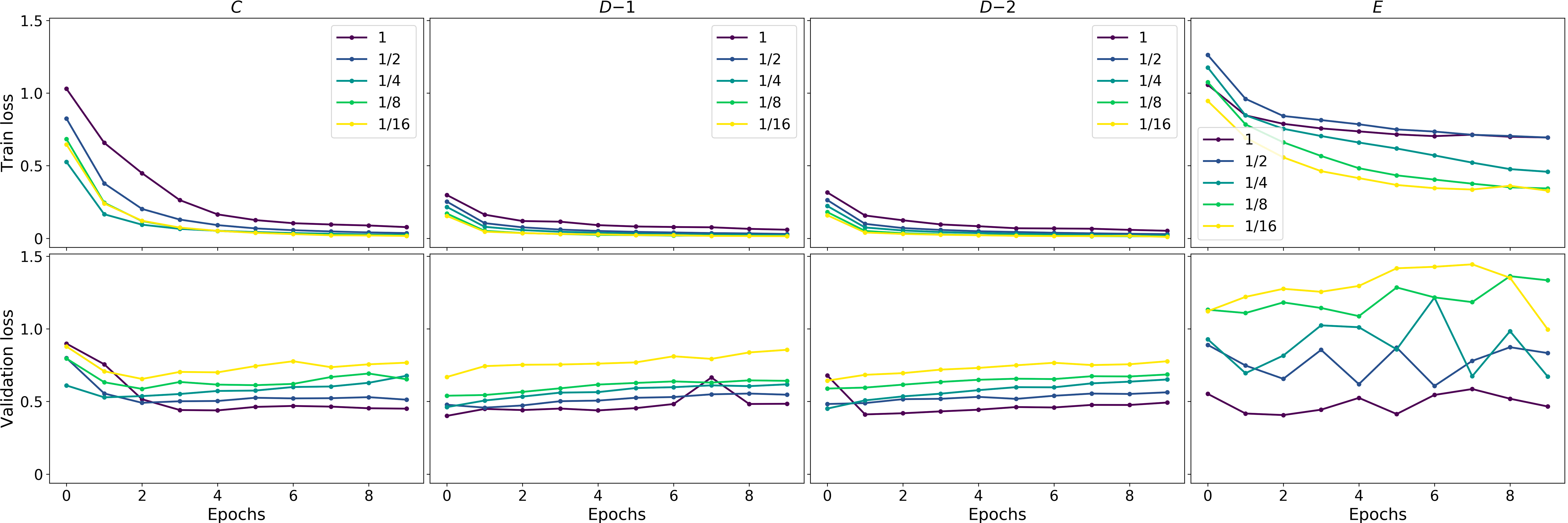}
    \caption{Learning progress of our models trained with different fractions of real-world data. Note that models \emph{D-1} and \emph{D-2} pre-trained on our synthetic dataset converge faster and remain robust to train dataset reduction.}
    \label{fig:results_learning_curves}
    \end{center}
\end{minipage} 
 \vspace*{-0.5cm}
\end{figure}

\endgroup
\section{Conclusion}
\label{sec:conclusion}

We have developed a pipeline for producing realistic synthetic data for the remote sensing domain. Using our pipeline, we have modeled the emergency mapping scenario and created 3D scenes and change detection image datasets of two real-world areas in California, USA. Results of the evaluation of deep learning models trained on our synthetic datasets indicate that synthetic data can be efficiently used to improve performance and robustness of data-driven models in real-world resource-poor remote sensing applications. We could further increase overall computational efficiency thanks to sparse CNNs \cite{3DCNN2018}, detection accuracy by using approaches to utilizing multi-modal data \cite{Multispectral2018}, imbalanced classification \cite{Imbalance2019,Imbalanced2015} and a loss, tailored for change detection in sequences of events \cite{Vehicle2017,EnsemblesDetectors2015}.



\bibliographystyle{splncs04}
\bibliography{references}

\begin{thebibliography}{10}
\providecommand{\url}[1]{\texttt{#1}}
\providecommand{\urlprefix}{URL }
\providecommand{\doi}[1]{https://doi.org/#1}

\bibitem{cryengine}
Cryengine. \url{https://www.cryengine.com}, accessed: 2019-01-30

\bibitem{cityengine}
Esri cityengine.
  \url{https://www.esri.com/en-us/arcgis/products/esri-cityengine/overview},
  accessed: 2019-01-30

\bibitem{OSM2xp}
Osm2xp. \url{https://wiki.openstreetmap.org/wiki/Osm2xp}, accessed: 2019-01-30

\bibitem{unity3d}
Unity. \url{https://unity3d.com}, accessed: 2019-01-30

\bibitem{ue4}
Unreal engine 4.
  \url{https://www.unrealengine.com/en-US/what-is-unreal-engine-4}, accessed:
  2019-01-30

\bibitem{WorldMachine}
World machine. \url{http://www.world-machine.com/}, accessed: 2019-01-30

\bibitem{alcantarilla2018street}
Alcantarilla, P.F., Stent, S., Ros, G., Arroyo, R., Gherardi, R.: Street-view
  change detection with deconvolutional networks. Autonomous Robots
  \textbf{42}(7),  1301--1322 (2018)

\bibitem{anniballe2018earthquake}
Anniballe, R., Noto, F., Scalia, T., Bignami, C., Stramondo, S., Chini, M.,
  Pierdicca, N.: Earthquake damage mapping: An overall assessment of ground
  surveys and vhr image change detection after l'aquila 2009 earthquake. Remote
  Sensing of Environment  \textbf{210},  166--178 (2018)

\bibitem{EnsemblesDetectors2015}
Artemov, A., Burnaev, E.: Ensembles of detectors for online detection of
  transient changes. In: Proc. SPIE. vol.~9875, pp. 9875 -- 9875 -- 5 (2015),
  \url{https://doi.org/10.1117/12.2228369}

\bibitem{babenko2014neural}
Babenko, A., Slesarev, A., Chigorin, A., Lempitsky, V.: Neural codes for image
  retrieval. In: European conference on computer vision. pp. 584--599. Springer
  (2014)

\bibitem{bai2018towards}
Bai, Y., Mas, E., Koshimura, S.: Towards operational satellite-based
  damage-mapping using u-net convolutional network: A case study of 2011 tohoku
  earthquake-tsunami. Remote Sensing  \textbf{10}(10), ~1626 (2018)

\bibitem{bourdis2011constrained}
Bourdis, N., Denis, M., Sahbi, H.: Constrained optical flow for aerial image
  change detection. In: 2011 IEEE International Geoscience and Remote Sensing
  Symposium (IGARSS). pp. 4176--4179 (2011)

\bibitem{bruzzone2014review}
Bruzzone, L., Demir, B.: A review of modern approaches to classification of
  remote sensing data. In: Land Use and Land Cover Mapping in Europe, pp.
  127--143. Springer (2014)

\bibitem{Multispectral2018}
Burnaev, E., Cichocki, A., Osin, V.: Fast multispectral deep fusion networks.
  Bull. Pol. Ac.: Tech.  \textbf{66}(4),  875--880 (2018)

\bibitem{Imbalanced2015}
Burnaev, E., Erofeev, P., Papanov, A.: Influence of resampling on accuracy of
  imbalanced classification. In: Proc. SPIE. vol.~9875, pp. 9875 -- 9875 -- 5
  (2015), \url{https://doi.org/10.1117/12.2228523}

\bibitem{Vehicle2017}
Burnaev, E., Koptelov, I., Novikov, G., Khanipov, T.: Automatic construction of
  a recurrent neural network based classifier for vehicle passage detection.
  vol. 10341, pp. 10341 -- 10341 -- 6 (2017),
  \url{https://doi.org/10.1117/12.2268706}

\bibitem{buslaev2018fully}
Buslaev, A., Seferbekov, S., Iglovikov, V., Shvets, A.: Fully convolutional
  network for automatic road extraction from satellite imagery. CoRR,
  abs/1806.05182  (2018)

\bibitem{cai2015detecting}
Cai, S., Liu, D.: Detecting change dates from dense satellite time series using
  a sub-annual change detection algorithm. Remote Sensing  \textbf{7}(7),
  8705--8727 (2015)

\bibitem{daudt2018urban}
{Caye Daudt}, R., {Le Saux}, B., Boulch, A., Gousseau, Y.: Urban change
  detection for multispectral earth observation using convolutional neural
  networks. In: IEEE International Geoscience and Remote Sensing Symposium
  (IGARSS) (July 2018)

\bibitem{chen2015deepdriving}
Chen, C., Seff, A., Kornhauser, A., Xiao, J.: Deepdriving: Learning affordance
  for direct perception in autonomous driving. In: Proceedings of the IEEE
  International Conference on Computer Vision. pp. 2722--2730 (2015)

\bibitem{chen2017deeplab}
Chen, L.C., Papandreou, G., Kokkinos, I., Murphy, K., Yuille, A.L.: Deeplab:
  Semantic image segmentation with deep convolutional nets, atrous convolution,
  and fully connected crfs. IEEE transactions on pattern analysis and machine
  intelligence  \textbf{40}(4),  834--848 (2017)

\bibitem{cheng2017remote}
Cheng, G., Han, J., Lu, X.: Remote sensing image scene classification:
  benchmark and state of the art. Proceedings of the IEEE  \textbf{105}(10),
  1865--1883 (2017)

\bibitem{deng2009imagenet}
Deng, J., Dong, W., Socher, R., Li, L.J., Li, K., Fei-Fei, L.: Imagenet: A
  large-scale hierarchical image database. In: Computer Vision and Pattern
  Recognition, 2009. CVPR 2009. IEEE Conference on. pp. 248--255. Ieee (2009)

\bibitem{dosovitskiy2017carla}
Dosovitskiy, A., Ros, G., Codevilla, F., Lopez, A., Koltun, V.: Carla: An open
  urban driving simulator. In: Conference on Robot Learning. pp. 1--16 (2017)

\bibitem{el2016convolutional}
El~Amin, A.M., Liu, Q., Wang, Y.: Convolutional neural network features based
  change detection in satellite images. In: First International Workshop on
  Pattern Recognition. vol. 10011, p. 100110W. International Society for Optics
  and Photonics (2016)

\bibitem{el2017zoom}
El~Amin, A.M., Liu, Q., Wang, Y.: Zoom out cnns features for optical remote
  sensing change detection. In: Image, Vision and Computing (ICIVC), 2017 2nd
  International Conference on. pp. 812--817. IEEE (2017)

\bibitem{fujita2017damage}
Fujita, A., Sakurada, K., Imaizumi, T., Ito, R., Hikosaka, S., Nakamura, R.:
  Damage detection from aerial images via convolutional neural networks. In:
  IAPR International Conference on Machine Vision Applications, Nagoya, Japan.
  pp. 08--12 (2017)

\bibitem{Gaidon2018}
Gaidon, A., Lopez, A., Perronnin, F.: The reasonable effectiveness of synthetic
  visual data. International Journal of Computer Vision  \textbf{126}(9),
  899--901 (Sep 2018). \doi{10.1007/s11263-018-1108-0},
  \url{https://doi.org/10.1007/s11263-018-1108-0}

\bibitem{gaidon2016virtual}
Gaidon, A., Wang, Q., Cabon, Y., Vig, E.: Virtual worlds as proxy for
  multi-object tracking analysis. In: Proceedings of the IEEE conference on
  computer vision and pattern recognition. pp. 4340--4349 (2016)

\bibitem{goodenough2012dirsig}
Goodenough, A.A., Brown, S.D.: Dirsig 5: core design and implementation. In:
  Algorithms and Technologies for Multispectral, Hyperspectral, and
  Ultraspectral Imagery XVIII. vol.~8390, p. 83900H. International Society for
  Optics and Photonics (2012)

\bibitem{goward2001landsat}
Goward, S.N., Masek, J.G., Williams, D.L., Irons, J.R., Thompson, R.: The
  landsat 7 mission: Terrestrial research and applications for the 21st
  century. Remote Sensing of Environment  \textbf{78}(1-2),  3--12 (2001)

\bibitem{gu2017change}
Gu, W., Lv, Z., Hao, M.: Change detection method for remote sensing images
  based on an improved markov random field. Multimedia Tools and Applications
  \textbf{76}(17),  17719--17734 (2017)

\bibitem{MedTransfer}
Haarburger, C., Langenberg, P., Truhn, D., Schneider, H., Thüring, J.,
  Schrading, S., Kuhl, C.K., Merhof, D.: Transfer learning for breast cancer
  malignancy classification based on dynamic contrast-enhanced mr images.
  Bildverarbeitung für die Medizin  (2018)

\bibitem{he2017mask}
He, K., Gkioxari, G., Doll{\'a}r, P., Girshick, R.: Mask r-cnn. In: Computer
  Vision (ICCV), 2017 IEEE International Conference on. pp. 2980--2988. IEEE
  (2017)

\bibitem{he2016deep}
He, K., Zhang, X., Ren, S., Sun, J.: Deep residual learning for image
  recognition. In: Proceedings of the IEEE conference on computer vision and
  pattern recognition. pp. 770--778 (2016)

\bibitem{huang2015automatic}
Huang, L., Fang, Y., Zuo, X., Yu, X.: Automatic change detection method of
  multitemporal remote sensing images based on 2d-otsu algorithm improved by
  firefly algorithm. Journal of Sensors  \textbf{2015} (2015)

\bibitem{iglovikov2018ternausnet}
Iglovikov, V., Shvets, A.: Ternausnet: U-net with vgg11 encoder pre-trained on
  imagenet for image segmentation. arXiv preprint arXiv:1801.05746  (2018)

\bibitem{ChangeDetectionRS2019}
Ignatiev, V., Trekin, A., Lobachev, V., Potapov, G., Burnaev, E.: Targeted
  change detection in remote sensing images. vol. 110412H (2019),
  \url{https://doi.org/10.1117/12.2523141}

\bibitem{jabari2016building}
Jabari, S., Zhang, Y.: Building change detection using multi-sensor and
  multi-view-angle imagery. In: IOP Conference Series: Earth and Environmental
  Science. vol.~34, p. 012018. IOP Publishing (2016)

\bibitem{jianya2008review}
Jianya, G., Haigang, S., Guorui, M., Qiming, Z.: A review of multi-temporal
  remote sensing data change detection algorithms. The International Archives
  of the Photogrammetry, Remote Sensing and Spatial Information Sciences
  \textbf{37}(B7),  757--762 (2008)

\bibitem{kemker2017deep}
Kemker, R., Kanan, C.: Deep neural networks for semantic segmentation of
  multispectral remote sensing imagery. CoRR, vol. abs/1703.06452  (2017)

\bibitem{kingma:adam}
Kingma, D.P., Ba, J.: Adam: A method for stochastic optimization. In:
  International Conference on Learning Representations (ICLR) (2015)

\bibitem{lee2018explainable}
Lee, H., Yune, S., Mansouri, M., Kim, M., Tajmir, S.H., Guerrier, C.E., Ebert,
  S.A., Pomerantz, S.R., Romero, J.M., Kamalian, S., et~al.: An explainable
  deep-learning algorithm for the detection of acute intracranial haemorrhage
  from small datasets. Nature Biomedical Engineering p.~1 (2018)

\bibitem{lin2017feature}
Lin, T.Y., Doll{\'a}r, P., Girshick, R., He, K., Hariharan, B., Belongie, S.:
  Feature pyramid networks for object detection. In: CVPR. vol.~1, p.~4 (2017)

\bibitem{lin2018focal}
Lin, T.Y., Goyal, P., Girshick, R., He, K., Doll{\'a}r, P.: Focal loss for
  dense object detection. IEEE transactions on pattern analysis and machine
  intelligence  (2018)

\bibitem{muller2018sim4cv}
M{\"u}ller, M., Casser, V., Lahoud, J., Smith, N., Ghanem, B.: Sim4cv: A
  photo-realistic simulator for computer vision applications. International
  Journal of Computer Vision pp. 1--18 (2018)

\bibitem{3DCNN2018}
Notchenko, A., Kapushev, Y., Burnaev, E.: Large-scale shape retrieval with
  sparse 3d convolutional neural networks. In: van~der Aalst, W.M., Ignatov,
  D., Khachay, M., et~al. (eds.) Analysis of Images, Social Networks and Texts.
  pp. 245--254. Springer International Publishing, Cham (2018)

\bibitem{novikov2018satellite}
Novikov, G., Trekin, A., Potapov, G., Ignatiev, V., Burnaev, E.: Satellite
  imagery analysis for operational damage assessment in emergency situations.
  In: International Conference on Business Information Systems. pp. 347--358.
  Springer (2018)

\bibitem{qiu2016unrealcv}
Qiu, W., Yuille, A.: Unrealcv: Connecting computer vision to unreal engine. In:
  European Conference on Computer Vision. pp. 909--916. Springer (2016)

\bibitem{richter2016playing}
Richter, S.R., Vineet, V., Roth, S., Koltun, V.: Playing for data: Ground truth
  from computer games. In: European Conference on Computer Vision. pp.
  102--118. Springer (2016)

\bibitem{ronneberger2015u}
Ronneberger, O., Fischer, P., Brox, T.: U-net: Convolutional networks for
  biomedical image segmentation. In: International Conference on Medical image
  computing and computer-assisted intervention. pp. 234--241. Springer (2015)

\bibitem{ImageNet}
Russakovsky, O., Deng, J., Su, H., Krause, J., Satheesh, S., Ma, S., Huang, Z.,
  Karpathy, A., Khosla, A., Bernstein, M., Berg, A.C., Fei-Fei, L.: Imagenet
  large scale visual recognition challenge. In: International Journal of
  Computer Vision (2015)

\bibitem{saha2018unsupervised}
Saha, S., Bovolo, F., Brurzone, L.: Unsupervised multiple-change detection in
  vhr optical images using deep features. In: IGARSS 2018-2018 IEEE
  International Geoscience and Remote Sensing Symposium. pp. 1902--1905. IEEE
  (2018)

\bibitem{shah2018airsim}
Shah, S., Dey, D., Lovett, C., Kapoor, A.: Airsim: High-fidelity visual and
  physical simulation for autonomous vehicles. In: Field and service robotics.
  pp. 621--635. Springer (2018)

\bibitem{sharif2014cnn}
Sharif~Razavian, A., Azizpour, H., Sullivan, J., Carlsson, S.: Cnn features
  off-the-shelf: an astounding baseline for recognition. In: Proceedings of the
  IEEE conference on computer vision and pattern recognition workshops. pp.
  806--813 (2014)

\bibitem{Imbalance2019}
Smolyakov, D., Korotin, A., Erofeev, P., Papanov, A., Burnaev, E.:
  Meta-learning for resampling recommendation systems. vol. 11041 (2019),
  \url{https://doi.org/10.1117/12.2523103}

\bibitem{tewkesbury2015critical}
Tewkesbury, A.P., Comber, A.J., Tate, N.J., Lamb, A., Fisher, P.F.: A critical
  synthesis of remotely sensed optical image change detection techniques.
  Remote Sensing of Environment  \textbf{160},  1--14 (2015)

\bibitem{vakalopoulou2015simultaneous}
Vakalopoulou, M., Karantzalos, K., Komodakis, N., Paragios, N.: Simultaneous
  registration and change detection in multitemporal, very high resolution
  remote sensing data. In: Proceedings of the IEEE Conference on Computer
  Vision and Pattern Recognition Workshops. pp. 61--69 (2015)

\bibitem{van2015off}
Van~Ginneken, B., Setio, A.A., Jacobs, C., Ciompi, F.: Off-the-shelf
  convolutional neural network features for pulmonary nodule detection in
  computed tomography scans. In: Biomedical Imaging (ISBI), 2015 IEEE 12th
  International Symposium on. pp. 286--289. IEEE (2015)

\bibitem{vittek2014land}
Vittek, M., Brink, A., Donnay, F., Simonetti, D., Descl{\'e}e, B.: Land cover
  change monitoring using landsat mss/tm satellite image data over west africa
  between 1975 and 1990. Remote Sensing  \textbf{6}(1),  658--676 (2014)

\bibitem{wang2017image}
Wang, B., Choi, J., Choi, S., Lee, S., Wu, P., Gao, Y.: Image fusion-based land
  cover change detection using multi-temporal high-resolution satellite images.
  Remote Sensing  \textbf{9}(8), ~804 (2017)

\bibitem{wiratama2018dual}
Wiratama, W., Lee, J., Park, S.E., Sim, D.: Dual-dense convolution network for
  change detection of high-resolution panchromatic imagery. Applied Sciences
  \textbf{8}(10), ~1785 (2018)

\bibitem{yao2010boosting}
Yao, Y., Doretto, G.: Boosting for transfer learning with multiple sources. In:
  Computer vision and pattern recognition (CVPR), 2010 IEEE conference on. pp.
  1855--1862. IEEE (2010)

\bibitem{yosinski2014transferable}
Yosinski, J., Clune, J., Bengio, Y., Lipson, H.: How transferable are features
  in deep neural networks? In: Advances in neural information processing
  systems. pp. 3320--3328 (2014)

\bibitem{yu2017change}
Yu, H., Yang, W., Hua, G., Ru, H., Huang, P.: Change detection using high
  resolution remote sensing images based on active learning and markov random
  fields. Remote Sensing  \textbf{9}(12), ~1233 (2017)

\bibitem{zhang2016deep}
Zhang, L., Zhang, L., Du, B.: Deep learning for remote sensing data: A
  technical tutorial on the state of the art. IEEE Geoscience and Remote
  Sensing Magazine  \textbf{4}(2),  22--40 (2016)

\bibitem{zhang2018perceptual}
Zhang, R., Isola, P., Efros, A.A., Shechtman, E., Wang, O.: The unreasonable
  effectiveness of deep features as a perceptual metric. In: CVPR (2018)

\end{thebibliography}

\end{document}